\documentclass[letterpaper, 10 pt, conference]{ieeeconf}  

\usepackage{amsmath} 
\usepackage{amssymb}  
\usepackage{dsfont}
\usepackage{graphicx}

\usepackage{psfrag,graphicx,epsfig}
\usepackage{epstopdf}
\usepackage{xspace}
\usepackage{subfig}
\usepackage{float}
\usepackage{placeins}
\usepackage{multirow}
\usepackage{pgf,tikz}
\usepackage{nowidow}
\usepackage{lineno}
\usepackage{xcolor}

\usepackage{siunitx}
\usepackage{color}
\usepackage{flushend}
\usepackage{lineno}
\usepackage{algorithmic}
\usepackage[vlined,linesnumbered,boxruled]{algorithm2e}

\usepackage{tikz} 
\usepackage{tikz-3dplot}

\def\radius{6}
\def\Alpha{20}
\def\Beta{20}
\def\Gamma{20}
\def\Gammazero{25}

\usepackage{caption}
\usepackage{booktabs}
\usepackage{varwidth}

\usepackage{wrapfig}

\usepackage{subfig}

\title{\LARGE \bf
Semiparametrical Gaussian Processes Learning of Forward Dynamical Models for Navigating in a Circular Maze
}

\author{Diego Romeres$^{1}$, Devesh K. Jha$^{1}$, Alberto DallaLibera$^{2}$, Bill Yerazunis$^{1}$ and Daniel Nikovski$^{1}$
\thanks{$^1$Diego Romeres, Devesh K. Jha, Bill Yerazunis and Daniel Nikovski are with Mitsubishi Electric Research Laboratories, Cambridge, MA 02139. Email--{\tt\small {\{romeres,jha, yerazunis, nikovski\}}@merl.com.}
$^2$ Alberto D. Libera is with Dept. of Information  Engineering, University of Padova, Via Gradenigo 6/b, 35131, Padova, Italy Email--{\tt \small{dallaliber@dei.unipd.it}}}
}

\begin{document}

\maketitle
\thispagestyle{empty}
\pagestyle{empty}

\begin{abstract}
This paper presents a problem of model learning for the purpose of learning how to navigate a ball to a goal state in a circular maze environment with two degrees of freedom. The motion of the ball in the maze environment is influenced by several non-linear effects such as dry friction and contacts, which are difficult to model physically. We propose a semiparametric model to estimate the motion dynamics of the ball based on Gaussian Process Regression equipped with basis functions obtained from physics first principles. The accuracy of this semiparametric model is shown not only in estimation but also in prediction at n-steps ahead and its compared with standard algorithms for model learning. The learned model is then used in a trajectory optimization algorithm to compute ball trajectories. We propose the system presented in the paper as a benchmark problem for reinforcement and robot learning, for its interesting and challenging dynamics and its relative ease of reproducibility.  
\end{abstract}
%
%
%
\section{Introduction}
\label{sec:Introduction}
One challenge of learning in physical systems is that data are expensive to obtain, and system dynamics are affected by non-linear and discontinuous phenomena such as dry friction, contact, hysteresis, etc~\cite{GS02}. It is, in general, very difficult to accurately model and/or learn the effect of these non-linear phenomena. Our work is motivated by learning the dynamics of physical systems that exhibit (highly) non-linear and discontinuous dynamics for the purpose of precise optimizing control. Deriving accurate models from first principles of physics could be extremely challenging, and in its turn, learning a model solely from data could be prohibitively expensive and inaccurate in sparsely sampled regions of state-space. For this reason,~semiparametric models which combine physical analytical models and data-driven techniques might be advantageous~\cite{kloss2017combining,fazeli2017learning, ajay2018augmenting}.

\begin{figure}[t]
  \centering
  \includegraphics[width=0.65\columnwidth]{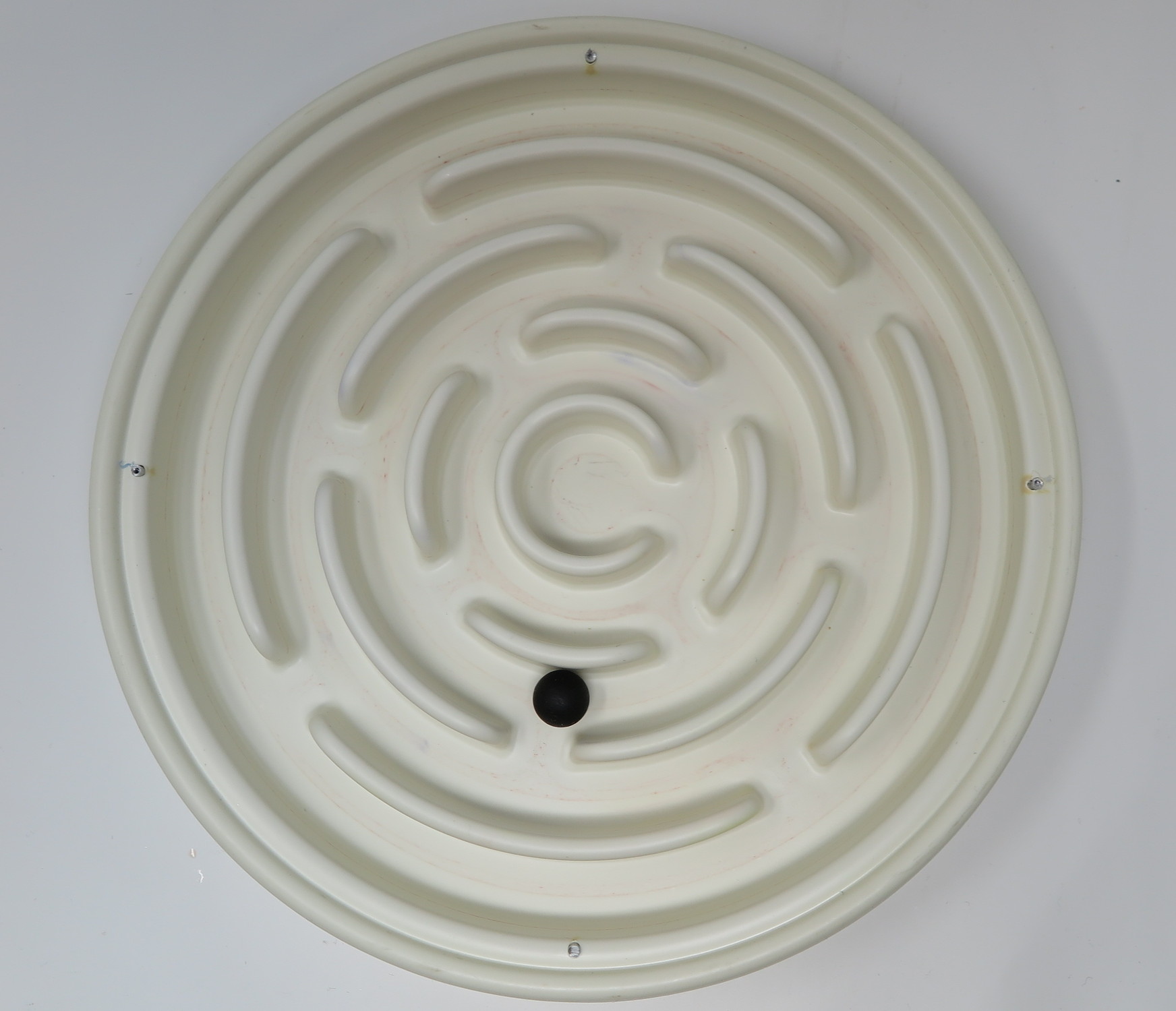}
  \caption{Circular Maze Environment (CME) with the ball}
  \label{fig:maze_image}
  \vspace{-2.2mm}
\end{figure}

This paper focuses on learning a dynamical model for the purpose of controlling the motion of a ball in a circular maze environment (see Figure~\ref{fig:maze_image}). The motion of the ball in the maze environment is significantly influenced by viscous and dry friction, contacts with the walls of the maze, and delays in sensing and actuation. Furthermore, the geometry of the maze constrains the motion of the ball to be discontinuous in the vicinity of each of the openings in each of the rings (see Figure~\ref{fig:maze_image}). As a result, it exhibits rich, highly non-linear dynamics. In this paper, we present results for model learning of the ball dynamics in a circular maze environment (CME), and initial results for trajectory optimization using~the learned models in a model-based reinforcement learning (MBRL)~setting.

\textbf{Related Work.} 
RL has seen explosive growth in recent years with several groundbreaking results in the areas of video games~\cite{silver2017, mnih2015}, robotics~\cite{levine2016}, etc. RL algorithms can be broadly classified into model-free and model-based~\cite{SB98, KBP13}. In the former class, the policy is directly inferred using either a policy or a value-based method~\cite{DNP13, ASM08}. In the latter, a model for the forward dynamics of the system is learned, and is then used to compute a policy to solve the desired task~\cite{DFR15, DR11, LK13}. Model learning could result in better generalization across several different tasks. However, learning global dynamics for nonlinear systems is generally considered a challenging problem. The use of Gaussian processes for model learning followed by a suitable controller design has been studied in the literature. Some of the related works which are closely related to the approach presented in this paper can be found in~\cite{DFR15, boedecker2014approximate, kamthe2017data, lee2017gp}. 
Apart from these, maze environments have been used to study RL algorithms ~\cite{St-Atkeson-06, zucker2012, bentivegna2001}. The proposed CME presents a more complex geometry, which results in significantly more convoluted motion dynamics of the ball, as will be presented in the later sections. Furthermore, the action space in the circular maze system (CMS) is continuous, whereas the action spaces considered in the earlier work with mazes~\cite{St-Atkeson-06, zucker2012, bentivegna2001} were discrete. In \cite{2018arXiv180904720V}, a deep RL-based end-to-end learning for the same CME is presented. The idea in this paper is to learn policies in simulation and transfer to the real system after minimum fine-tuning. 

\textbf{Contributions.} 
Our contributions are the following. First,  we present a semiparametric modeling technique \cite{romeres2016onlineIcub,wu2012semi,ICRA2010NguyenTuong_62320} which combines the physical model of the system with data-driven Gaussian Process Regression \cite{Rasmussen} to learn the forward dynamics of the ball in a novel, real physical system. To the best of our knowledge, semi-parametric Gaussian process regression has only been used to learn inverse dynamics previously~\cite{2018arXiv180905074R,ICRA2010NguyenTuong_62320,romeres2016onlineIcub,Nguyen-Tuong2011}. Second, continuous-time forward dynamics are learned, and not discrete-time state transition functions. We estimate the ball acceleration signal, and then integrate it forward in time to predict the state; we do not directly estimate the successor state. Third, performance of the modeling choices are shown in long-horizon predictive capabilities (known to be a hard task for the compounding error effect) and compared with several other techniques to highlight its accuracy. Rollout accuracy is fundamental to obtain successful optimal control algorithms that use known dynamics. Fourth, we illustrate the control performance in the CME using the learned models and the Iterative Linear Quadratic Gaussian (iLQG) algorithm~\cite{todorov2005, tassa2012}. We make simplifying assumptions to decompose the full goal-directed navigation into a sequence of sub-goals in each of the rings. We show that using the learned models, we can control the ball between arbitrary points in the maze environment, thus also navigating to the goal ring. In the remainder of the paper, CMS refers to the full system comprising of the CME and the tip-tilt platform.
\begin{figure}[t]
  \centering
  \includegraphics[width=.58\columnwidth]{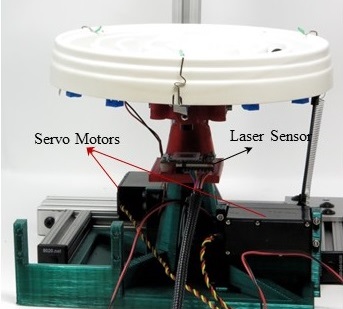}
  \caption{Circular Maze System (CMS) setup with the maze housed on the tip-tilt stage. Camera is mounted on top of the tip-tilt stage.}
  \label{fig:system_setup_images}
  \vspace{-4mm}
\end{figure}

\section{Experimental Configuration}
\label{sec:exp_config}
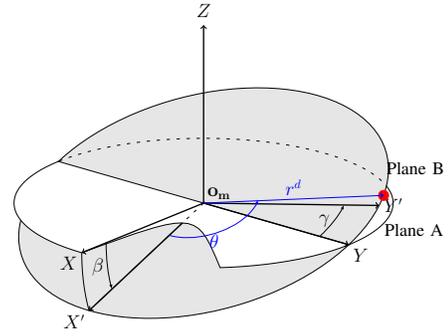
\begin{figure}[t]
\centering
\tdplotsetmaincoords{70}{150}

\tikzset{
    axis/.style={very thick, fill=white, -*, shorten >=-3pt},
    axis'/.style={very thick, ->},
    axis''/.style={thick, -},
    Axis/.style={very thick, fill=black, -*, shorten >=-3pt},
    behind lines/.style={loosely dashed}
}
\scalebox{0.420}{
\begin{tikzpicture}[tdplot_main_coords]

\begin{scope}[scale=\radius, every path/.style={tdplot_rotated_coords}]

\tdplotsetrotatedcoords{\Alpha}{0}{0}
\draw  [fill=white] 
    plot    [domain=90:270, samples=100] (cos \x, sin \x) -- cycle;

\tdplotsetrotatedcoords{\Alpha}{\Beta}{\Gamma}
\draw  [fill=gray!20] 
    plot    [domain=0:360, samples=100] (cos \x, sin \x);

\begin{scope}
\tdplotsetrotatedcoords{\Alpha}{\Beta}{\Gamma}
\path  [clip] 
    plot    [domain=0:360, samples=100] (cos \x, sin \x);

\tdplotsetrotatedcoords{\Alpha}{0}{0}
\draw  [behind lines] 
    plot    [domain=90:270, samples=100] (cos \x, sin \x);
\end{scope}


\tdplotsetrotatedcoords{\Alpha-90}{-90}{90}
\draw [thick, domain=0:\Beta] 
    (0,0) -- plot (cos \x, sin \x) -- cycle;


\tdplotsetrotatedcoords{\Alpha}{\Beta}{0}
\draw[axis'] (0,0,0) -- (1,0,0) node (X'-axis) [below left] {\LARGE$X'$};
\node[above right] (0,0,0){\LARGE$\mathbf{o_m}$};

\tdplotsetrotatedcoords{\Alpha}{0}{0}
\draw  [fill=white] 
    plot [domain=-90:0, samples=100] (cos \x, sin \x) 
    .. controls (0.125*cos 22.5, 0.125*sin 22.5) ..
    (cos 45, sin 45) -- plot    [domain=45:90, samples=100] (cos \x, sin \x) 
    -- cycle;

\begin{scope}
\tdplotsetrotatedcoords{\Alpha}{0}{0}
\path  [clip] 
    plot [domain=-90:0, samples=100] (cos \x, sin \x) 
    .. controls (0.125*cos 22.5, 0.125*sin 22.5) ..
    (cos 45, sin 45) -- plot    [domain=45:90, samples=100] (cos \x, sin \x) 
    -- cycle;

\tdplotsetrotatedcoords{\Alpha-90}{-90}{90}
\draw [behind lines] (0,0) -- (cos \Beta, sin \Beta);


\end{scope}

\tdplotsetrotatedcoords{0}{0}{0}
\draw[axis'] (0,0,0) -- (0,0,1) node [above=0.1cm] {\LARGE$Z$};

\tdplotsetrotatedcoords{\Alpha}{0}{0}
\draw[axis'] (0,0,0) -- (1,0,0) node [below left] {\LARGE$X$};

\tdplotsetrotatedcoords{\Alpha}{\Beta}{\Gamma}
\draw[axis'] (0,0,0) -- (0,1,0) node [right] {\LARGE$Y'$};

\tdplotsetrotatedcoords{\Alpha}{\Beta}{0}
\draw[axis'] (0,0,0) -- (0,1,0) node [below right] {\LARGE $Y$};
\tdplotsetrotatedcoords{\Alpha}{\Beta}{\Gammazero}
\filldraw[fill=red,draw=red] (0,1,0) circle (.75pt);
\draw[thick,->,blue](0,0,0)--(0,1,0) node [above] at (0, 0.5,0) {\LARGE $r^d$};
\node [above] at (0,1.2,0.1) {\LARGE Plane B};
\node [above] at (0,1.1,-0.3) {\LARGE Plane A};
\tdplotsetrotatedcoords{\Alpha}{\Beta}{\Gammazero}
\tdplotdrawarc[-latex,blue]{(0,0,0)}{0.3}{0-\Gammazero}{92}{anchor=north east}{\LARGE$\theta$}
\tdplotsetrotatedcoords{0}{0}{0}
\tdplotsetrotatedcoords{\Alpha+90}{90}{0}
\tdplotdrawarc[-latex]{(0,0,0)}{0.8}{-90}{-90+\Beta}{left}{\LARGE $\beta$}
\tdplotsetrotatedcoords{\Alpha}{\Beta}{\Gamma}

\tdplotsetrotatedcoords{0}{0}{0}
\tdplotsetrotatedcoords{\Alpha+90}{90}{0}
\tdplotsetrotatedcoords{\Alpha}{\Beta}{\Gamma}
\tdplotdrawarc[-latex]{(0,0,0)}{0.8}{90-\Gamma}{90}{left}{\LARGE $\gamma$}

\end{scope}

\end{tikzpicture}}
\caption{Coordinate reference frames for the maze environment on the tip-tilt platform. Angles $\beta$ and $\gamma$ represent the orientation of the tip-tilt stage. The ball is shown in red.}
\label{fig:Coordinate_frame}
\vspace{-6 mm}
\end{figure}

The electromechanical tip-tilt system is composed of the maze, a maze attachment plate, two gimbals, two RC servos, a 2D laser tilt sensor, a base plate, and and Arduino-based control processor (see Figure~\ref{fig:system_setup_images}).
Camera input is provided by an Intel RealSense camera on an overhead mount (see Figure~\ref{fig:system_setup_images}). 
All the communication from the camera through the image processing system, the laser sensor for measuring the inclination of the tip-tilt platform, and the servo motors driving the tip-tilt stage is done using the Robot Operating System (ROS)~\cite{ROS09} at 30 \si{\hertz}.\\
The maze itself is a commercially available toy purchased from Amazon.com (see Figure~\ref{fig:maze_image}).
A 3D printed plate with two positioning setscrews and three spring clips fits tightly inside the base of the maze and provides attachment points for the servo link rods and the gimbals. The gimbals constrain the platform to have pure rotational motion around its origin $\mathbf{o_{m}}$. The latter, represents the Cartesian coordinates of the rotational fulcrum identified by the gimbals in the inertial frame $(\mathbf{o_{b}},\mathbf{x},\mathbf{y},\mathbf{z})$.\\
 Without loss of generality, we can assume that the origin of the inertial frame coincides with $\mathbf{o_{m}}$ (see Figure~\ref{fig:Coordinate_frame}). The configuration of the platform could be fully described by the rotation matrix between frame $(\mathbf{o_{b}},\mathbf{x},\mathbf{y},\mathbf{z})$ and the frame attached to the maze $(\mathbf{o_{m}},\mathbf{x'},\mathbf{y'},\mathbf{z'})$; this can be described in a more compact fashion using the yaw, pitch and roll angles. The maze base is constrained by servo link rods, attached to the maze base through Traxxis ball joints; the lower end of each link rod is connected to a bellcrank arm of a hobbyist-grade HS-805BB RC servo motor with PWM-based control. The motors are positioned in perpendicular directions, so that the transformations between $(\mathbf{o_{b}},\mathbf{x},\mathbf{y},\mathbf{z})$ and $(\mathbf{o_{m}},\mathbf{x'},\mathbf{y'},\mathbf{z'})$ is described only by $\gamma$ and $\beta$, the roll and pitch angles. The HS-805BB servos allow only position control. The control target of the motors, namely the input of our system, will be denoted by the vector $\mathbf{u}=(u^\beta,u^\gamma)$. The HS-805BB controller takes in the order of $200$\si{\milli\second} to move from an initial to a desired angle. 
Indirect measurements of $\gamma$ and $\beta$ are acquired using an inexpensive 5 milliwatt diode laser (Adafruit productid 1054), placed under the maze plate such that the direction of its optical axis is always equal to $\mathbf{z'}$. The laser projects downward onto the non-moving tilt position sensor, a $10\si{mm}\times10\si{mm}$ position sensitive diode (a FirstSensor DL100-7-PCBA3 position sensitive photodiode). This sensor emits a pair of voltages corresponding to the x and y component of the impinging light. The map from the laser sensor readings to $\gamma$ and $\beta$ is learned by calibrating the laser readings with a precision ($0.1^{\circ}$) clinometer. \\
 The radius of the maze is approximately $110mm$,  and the diameter of the ball is approximately $12.75mm$. The width of each of the openings that connect rings is around $16mm$. This suggests that the controller has to be fairly accurate in order to be able to control the ball in the CME.  The position of the ball is expressed in polar coordinates by $r$ and $\theta$, which represent the distance between $\mathbf{o_{m}}$ and the center of the ball, and the rotation around $\mathbf{z}'$ respectively. The position information is obtained using the camera and an off-the-shelf blob detection algorithm; angular velocity $\dot{\theta}$ is estimated using a Kalman filter.\\
The entire CMS except for the camera is connected to a computer via an Arduino Mega $2560$. The entire mechanical system with the exception of the bearings and servo motors were designed  in-house with OpenSCAD, with most of the non-catalog parts printed in InPLA polymer on an Lulzbot TAZ 3 printer.

\vspace{-3mm}
\section{Problem Formulation}
\label{sec:problem_formulation}
In this section, we describe the task in the proposed CMS and the assumptions considered in order to decompose the original problem into sub-problems with lower complexity. Throughout the paper, we assume that the relationship between observations and state is deterministic and known. The system is thus fully defined by the combination of the state $\mathbf{x}_k$ and the control inputs $\mathbf{u}_k$, and it evolves according to the dynamics $p(\mathbf{x}_{k+1}\vert \mathbf{x}_k, \mathbf{u}_k)$ which are composed of the ball dynamics in the maze and the tip-tilt platform dynamics. The goal is to learn an accurate model of the ball dynamics which can be used in a controller, $\pi(\mathbf{u}_k \vert \mathbf{x}_k)$, based on a trajectory optimization algorithm which allows the CMS to choose an action $\mathbf{u}_k$ given the state observation $\mathbf{x}_k$ to drive a ball from an initial condition to the target state.  
As a first simplification, we assume that the ball dynamics are independent of the radial dynamics in each of the individual rings -- i.e., we quantize the radius of the ball position into the 5 rings of the maze, denominated from 1 to 5 starting from the most outer ring toward the most inner one. 
The radius of a ring and therefore the radius of the ball position is approximated with $r_d$, the mean radius of the ring in which the ball is moving.
As discussed in Section~\ref{sec:exp_config}, the actuators for the tip-tilt platform allow only position control, and they experience significant delay (around 200 [\si{\milli\second}]) in attaining the desired angle. Thus, we include the orientation of the tip-tilt stage as part of the state for our dynamical system obtaining a seven-dimensional state representation for the system, i.e., $\mathbf{x}=(r^d,\theta,\dot{\theta},\beta,\dot{\beta},\gamma,\dot{\gamma})$. It is noted that the radius $r^d$ is a discrete variable, while the rest of the state variables are continuous. The input space for the system is the two dimensional vector $\mathbf{u}$ defined in Section~\ref{sec:exp_config}.\\
To further simplify the problem, we decompose the problem into macro and micro goals. While the macro goal of the ball is to be able to reach the goal ring, this goal can be composed of a sequence of micro goals. For example, from any initial position, the ball should reach one of the openings and then move to the next ring. This sequence of movements should be repeated until the ball reaches the goal ring. This decomposition is motivated by discontinuity in ball dynamics when the ball moves to an adjacent ring through one of the openings. Such a decomposition of long horizon planning problems, with sparse discontinuities, are common in the robot learning literature~\cite{kroemer2014, lioutikov2015, kroemer2015}. While the full long-horizon control problem could be challenging to solve using a MBRL algorithm, the sub-problems for each of the micro-goals could be solved by model learning followed by trajectory optimization. 
%
\section{Proposed Learning Approach}
\label{sec:mbrl_algorithm}
In this section, we describe the proposed solution to learn how to navigate a ball in the CME. First, we learn a Gaussian process-based model for movement in each ring. A full motion model for the entire maze is obtained by using an additional discrete model for the movement between rings. Second, the learned models are used to compute a controller using iLQG to learn a local policy to navigate in the maze. %
\subsection{Model Learning}
\label{subsec:model_learning}
\vspace{-1 mm}
We consider discrete-time systems:
\begin{equation}
\label{eq:discrete_dynamics_general}
    \mathbf{x}_{k+1} = f(\mathbf{x}_k, \mathbf{u}_k) + \mathbf{e}_k
\end{equation}
where $\mathbf{x}_k \in \mathbb{R}^7$ denotes the state, $\mathbf{u}_k \in \mathbb{R}^2$ the actions, and $\mathbf{e}_k$ is assumed to be a zero mean white Gaussian noise with diagonal covariance that represents the uncertainty about the state at the discrete time instant $k \in [1, \ldots, T]$. The transition function $f$ maps the state-action pair to the successor state, which is assumed to evolve smoothly over time. Given the quantization of the radius introduced in Section \ref{sec:problem_formulation}, we do not use the measurements of the radius of the ball's position. Even though this leads to information loss, preventing the possibility of modeling complex dynamics behaviours such as collisions, it reduces modeling complexity. The radial dynamics in the model will be incorporated in future works. \\
The state $\mathbf{x}$ for the CMS can be decomposed into two subsets: the ball state $\mathbf{x}^{\textit{ball}} = [r^d, \theta, \dot{\theta}]$, and the tip-tilt platform state $\mathbf{x}^{\textit{tt}} = [\beta, \dot{\beta},\gamma, \dot{\gamma}]$. This decomposition is convenient, because it is reasonable to assume that the tip-tilt platform dynamics are independent of the ball dynamics which is
\begin{equation}
\label{eq:ttstage_dynamics_independent}
p(\mathbf{x}^{\textit{tt}}_{k+1}\vert \mathbf{x}_{k}, \mathbf{u}_{k}) = p(\mathbf{x}^{\textit{tt}}_{k+1}\vert \mathbf{x}_{k}^{\textit{tt}}, \mathbf{u}_{k})
\end{equation}
In the proposed approach, we do not learn directly the static map \eqref{eq:discrete_dynamics_general}, but we instead learn the evolution of $\mathbf{\dot{x}_t}$, which is a second order continuous-time system (subscript $t$ denotes continuous time). This is different from earlier work presented in~\cite{DFR15, DR11} where the discrete-time map was learned. For our system, this reduces to the estimation of three ordinary differential equations for the acceleration functions of the positional states $[\theta, \beta, \gamma]$. This function can then be integrated forward in time to make predictions on future states. In our opinion, this choice has the advantage that $\mathbf{\dot{x}_t}$ is described by the underlying physical system, and this knowledge can be exploited more efficiently in learning dynamic models. Notice also that in discrete time the evolution of $\mathbf{x}^{\textit{ball}}$ given $\mathbf{x}_k$ is not independent of $\mathbf{u}_k$, since we need to know the input $\mathbf{u}_k$ until the next observation to make future predictions. In the continuous case, we have that $\mathbf{\dot{x}}^{ball}_t$ is independent from $\mathbf{u}_t$ given $\mathbf{x}_t$, $\mathbf{x}^{tt}_t$ and $\mathbf{\dot{x}}^{tt}_t$, which is
\begin{equation*}
p(\mathbf{\dot{x}}_t^{\textit{ball}}\vert \mathbf{x}_t, \mathbf{u}_t) = p(\mathbf{\dot{x}}_t^{\textit{ball}}\vert \mathbf{x}_t, \mathbf{\dot{x}}^{tt}_t)
\end{equation*}
The continuous time models are then discretized to compute the models used for control. In the following, we focus on learning acceleration models to be able to describe \eqref{eq:discrete_dynamics_general}.\\ 
First, we model the dynamics of the tip-tilt platform. Because of \eqref{eq:ttstage_dynamics_independent}, the tip-tilt stage dynamics \eqref{eq:discrete_dynamics_general} can be learned independently from the ball dynamics. The dynamics of 
$\beta$ and $\gamma$ have been shown to be modelled effectively from physical quantities with a linear second order continuous-time differential equation \cite{anjali2016implementation,campos2004optimal}. Therefore, the acceleration functions $\ddot{\beta}_t = f^\beta(\dot{\beta}_t,\beta_t, \gamma_t, u_t^\beta)$ and $\ddot{\gamma}_t = f^\gamma(\dot{\gamma}_t,\gamma_t, \beta_t, u_t^\gamma)$  have been estimated using standard linear system identification techniques \cite{Ljung:99} using Prediction Error Methods, resulting in two linear ARX models for each angle $\beta$ and $\gamma$.\\
Next, we focus on modelling the more complex ball dynamics. Additionally, we define a new state vector $\mathbf{\bar{x}}_t$ obtained by stacking together $\mathbf{x}_t$ and $\mathbf{\dot{x}}^{tt}_t$ 
\begin{equation}
\label{eq:continuous_dynamics_ball}
    \ddot{\theta}_t = f^{\textit{ball}}(\mathbf{x}_t, \mathbf{\dot{x}}^{tt}_t) + \mathbf{e}_t = f^{\textit{ball}}(\mathbf{\bar{x}}_t) + \mathbf{e}_t 
\end{equation} 
Most of the models introduced in this section are equipped with a set of unknown \textit{hyperparameters} that we estimate using the maximization of the marginal likelihood. Since all the variables are referred to time $t$, from now on we will neglect the dependencies on $t$.

\subsubsection{Physics Inspired Model}
\label{subsubsec:simplified_maze_model}
We are interested in determining an analytical expression of the forward dynamics~\eqref{eq:continuous_dynamics_ball}, and using it for data-driven modeling. It is noted that this model only describes the motion of the ball in a circular ring, and doesn't consider the tip-tilt platform. As described in Section~\ref{sec:problem_formulation}, the radial dynamics of the ball is ignored, and the analytical model will be computed for a given radius $r$ that is assumed to be constant. Using a standard Lagrangian approach~\cite{siciliano2010robotics,bullo2004,hollerbach2008model}, we obtain the following equations of motion (we skip a detailed derivation for brevity) :%
\begin{small}
\begin{align}
\ddot{\theta} = &\ddot{\beta}\sin(\gamma) - \dot{\beta}^2\sin(\theta)\cos(\gamma)^2\cos(\theta) - 2\dot{\beta}\dot{\gamma}\cos(\gamma)\cos(\theta)^2 \nonumber \\  &+2\dot{\beta}\dot{\gamma}\cos(\gamma) \nonumber +  \frac{1}{2}\dot{\gamma}^2\sin(2\theta)
  g\sin(\beta)\sin(\theta)/r\\
  &- g\sin(\gamma)\cos(\beta)\cos(\theta)/r 
\label{eqn:simplified_ball_dynamics}
\end{align}
\end{small}
Equation~\eqref{eqn:simplified_ball_dynamics} reflects the complex dynamics of ball movement due to the circular geometry of the maze when compared to relatively simple geometries considered earlier in~\cite{kroemer2014, lioutikov2015, kroemer2015}. We refer to this model as the Physical model, ``P".
While P in equation~\eqref{eqn:simplified_ball_dynamics} may be inaccurate for the CME due to making simplifying assumptions, it does provide a well-defined structure to the model that can be hard to identify in data-driven models in an efficient fashion.
This model structure is used and interpret each component of the right-hand side of equation \eqref{eqn:simplified_ball_dynamics} as basis functions of the feature space of a Bayesian linear model as defined in \cite[Chp. 2]{Rasmussen}:
\begin{small}
\begin{align}
\phi^\top(\mathbf{\bar{x}}) = &\left[ \ddot{\beta}\sin(\gamma), -\dot{\beta}^2\sin(\theta)\cos(\gamma)^2\cos(\theta), 2\dot{\beta}\dot{\gamma}\cos(\gamma) , \nonumber \right. \\
 & - 2\dot{\beta}\dot{\gamma}\cos(\gamma)\cos(\theta)^2,  0.5\dot{\gamma}^2\sin(2\theta)\sin(\beta)\sin(\theta)g/r, \nonumber \\
 & \left. - g\sin(\gamma)\cos(\beta)\cos(\theta)g/r \right]
\label{eq:physical_features}
\end{align}
\end{small}
The expression of the forward dynamics \eqref{eq:continuous_dynamics_ball} becomes
\begin{align}
\ddot{\theta} &= \phi^\top(\mathbf{\bar{x}}) \mathbf{w} + \mathbf{e}
\label{eq:PP_dynamics}
\end{align}
where $\mathbf{w} \sim {\cal N} (0, \Sigma_w)$ is a vector of weights modeled as a zero mean Gaussian prior with diagonal covariance matrix $\Sigma_w$. This model is denoted as Physics-Inspired model,~``PI".
\subsubsection{Nonparametric Model}
\label{subsubsec:nonparametric_model}
According to the Gaussian Process Regression framework \cite{Rasmussen}, the forward dynamics can be postulated in a data driven fashion as:
\begin{equation}
\ddot{\theta} = g(\mathbf{\bar{x}}) + \mathbf{e}
\label{eq:NP_dynamics}
\end{equation}
where $ g(\mathbf{\bar{x}})$ is a Gaussian process with zero mean and covariance function $k(\mathbf{\bar{x}},\mathbf{\bar{x}}'):= \mathbb{E}[g(\mathbf{\bar{x}}) g(\mathbf{\bar{x}}')]$ defined through a radial basis function (RBF) kernel.
This data-driven model class is known to have high flexibility in modelling functions and also high prediction performance, because it extrapolates the dynamics directly from the data without relying on any physical assumptions. 
However, this method can perform poorly in sparsely sampled regions of the state-space. This model is denoted as the Nonparametric model, ``NP".

\subsubsection{Semiparametric Model}
\label{subsubsec:simplified_maze_model}
Semiparametric models attempt to combine the benefits of both the physical and data-driven  model classes, which are good global behavior and accuracy, respectively. There can be several ways of combining the two methods and choosing the a priori distribution, as shown in \cite{romeres2016onlineIcub}, but one of the most effective has been shown to be:
\begin{equation}
\ddot{\theta} = \phi^\top(\mathbf{\bar{x}}) \mathbf{w} + g(\mathbf{\bar{x}}) + \mathbf{e}
\label{eq:SP_dynamics}
\end{equation}
where $\phi^\top(\mathbf{\bar{x}})$ are the basis function suggested by the physics and defined as \eqref{eq:physical_features}, $\mathbf{w}$ is a vector of parameters with a zero mean Gaussian prior and a diagonal covariance matrix $\Sigma_w$ and $g(\mathbf{\bar{x}})$ is a zero mean Gaussian process with RBF covariance function. The covariance of the whole process is $k(\mathbf{\bar{x}},\mathbf{\bar{x}}'):= \phi^\top(\mathbf{\bar{x}}) \Sigma_w \phi(\mathbf{\bar{x}}) + \mathbb{E}[g(\mathbf{\bar{x}}) g(\mathbf{\bar{x}}')]$.
This class of models is also known as semi-parametrical models, and it has been shown to be a successful technique in model learning for robotic applications in the recent literature \cite{romeres2016onlineIcub,wu2012semi,ICRA2010NguyenTuong_62320}. This model is denoted as Semiparametric model,~``SP''.
%
\begin{figure*}
\flushleft
\begin{minipage}{0.61\textwidth}
\centering
\captionsetup{type=table} 
\small
\setlength\tabcolsep{2.5pt} 
\begin{tabular}{ c c c c c c c c c c c c c c c}
      \toprule
      & \multicolumn{7}{c}{nRMSE Training} & \multicolumn{6}{c}{nRMSE Test} \\
      \midrule
      Ring & $N_{tr}$ & $c_{DS}$ & RF & P & PI & NP & SP &  $N_{test}$ & RF & P & PI & NP & SP \\
      \midrule
      1 & 3874 & 5 & 0.18 & 2.07 & 0.49 & 0.31 & 0.28 & 19368 & 0.49 & 1.97 & 0.48 & 0.38 & 0.35 \\
      2 & 4365 & 4 & 0.22 & 1.56 & 0.63 & 0.31 & 0.29 & 17459 & 0.59 & 1.57 & 0.63 & 0.45 & 0.43 \\
      3 & 4606 & 2 & 0.19 & 1.32 & 0.66 & 0.19 & 0.16 & 9213 & 0.53 & 1.35 & 0.65 & 0.42 & 0.40\\
      4 & 2814 & 1 & 0.15 & 1.10 & 0.79 & 0.02 & 0.01 & 2815 & 0.67 & 1.08 & 0.81 & 0.71 & 0.65\\
      \bottomrule
\end{tabular}
\caption{nRMSE performance in training and in test data in all the \\rings for all the presented models and a Random Forest model.}
\label{table:nRMSE_1step}
\end{minipage}
\begin{minipage}{0.38\textwidth}
\flushleft
\qquad \quad
\includegraphics[width=1.02\textwidth]{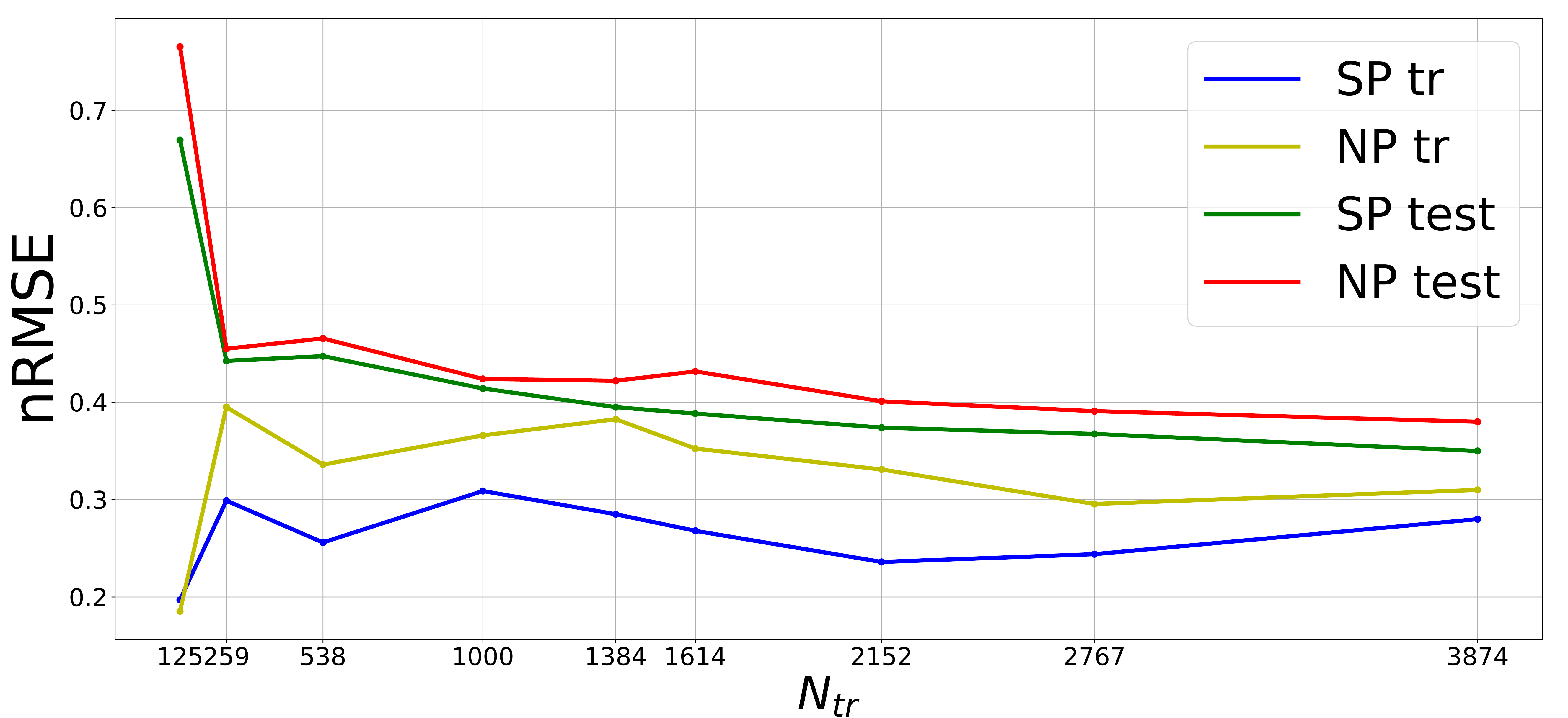}%
\vspace{-2mm}
\caption{Learning curves of SP and NP in training and test as the number of training data increases for ring 1.}
\label{fig:nRMSE_learning_curve}
\end{minipage}
\vspace{-5 mm}
\end{figure*}
\subsection{Radial Movement}
\label{subsec:radial_movement}
The model of the continuous movement of the ball in each of the rings was described in Section \ref{subsec:model_learning}. Next, we model the discrete movement of the ball between rings. To learn this movement, we first learn the spatial clusters where there is a transition in $r^d$ in the collected trajectories of the ball. Thus, we obtain two clusters in the vicinity of each opening, based on the sign of $\Delta r^d$. The radial transition between rings is then approximated by the vector joining the centroids of the two clusters on the two sides of an opening, and the control action $\mathbf{u}$ is approximated as the maximum signal whose projection on the horizontal plane is parallel to the direction of the radial transition vector.  
\vspace{-1 mm}
\subsection{Trajectory Optimization using iLQG}
\label{subsec:IterativeLQG}
In this section, we briefly describe the iLQG algorithm which is used for trajectory optimization. The discrete time dynamics $\mathbf{x}_{k+1} = f(\mathbf{x}_k, \mathbf{u}_k)$ and the cost function are used to compute local linear models and a quadratic cost function for the system along a trajectory. These linear models are then used to compute optimal control inputs and local gain matrices by iteratively solving the associated LQG problem. We skip the mathematical details for space limitations. For more details, interested readers are referred to~\cite{tassa2012}. 
%
%
%
\section{Experimental Results}
\label{sec:exerimental results}
\begin{figure}[h]
\includegraphics[width=\columnwidth]{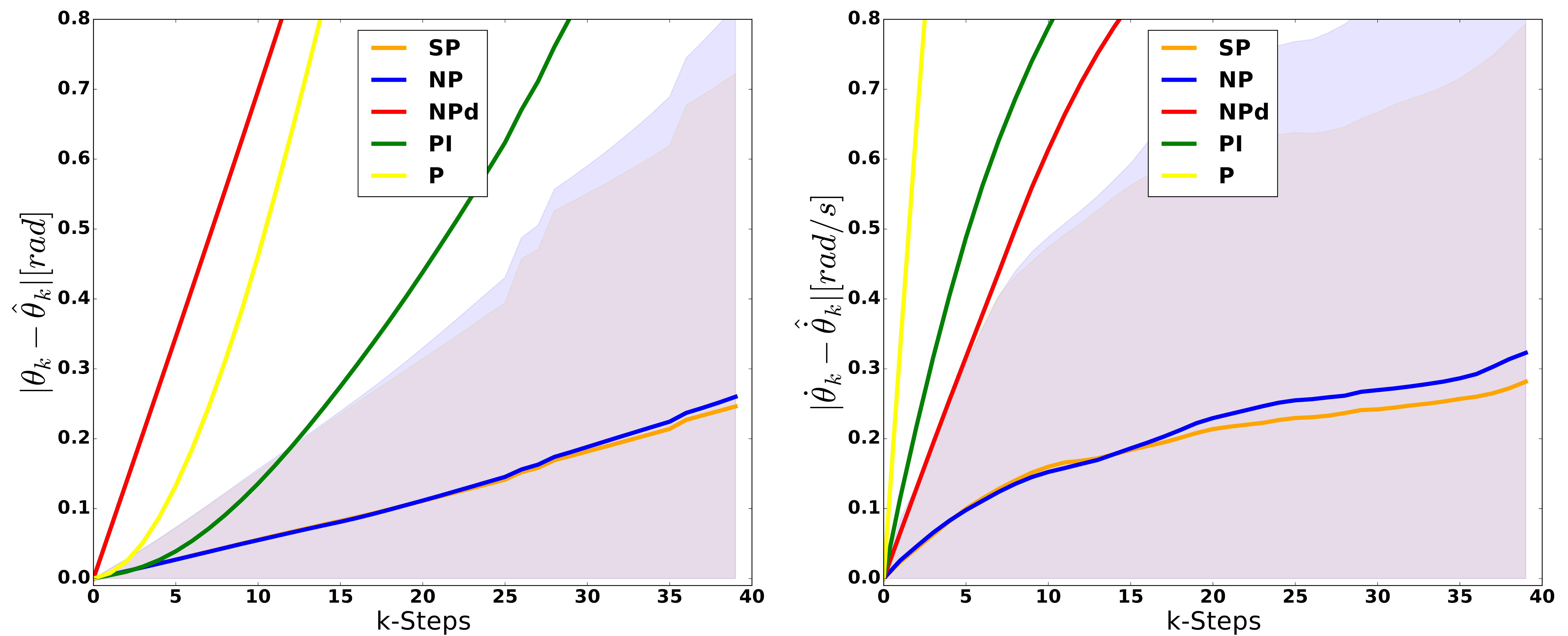}%
\caption{Rollout performance. The average absolute error in $\theta$ and $\dot{\theta}$ at k-steps ahead are shown for all the models, on the left and right-hand side, respectively.}\label{fig:rollout_performance}
\end{figure}

In this section, we will provide experimental results for the models presented in Section~\ref{subsec:model_learning}, and we will show their efficacy in the real system, CMS, by learning a policy as described in Section~\ref{subsec:IterativeLQG}.\\
Data collection consisted of around 50 minutes of operation on CMS, when control actions were applied at 30Hz, and measurements from the laser and camera were recorded. During data collection, control actions were generated as a sum of 50 sine and cosine waves with random sampled frequency between $[0,1.5]$ Hz and shift phases in $[0,2\pi]$.
The data set consists of the control actions matrix $\mathbf{U} \in \mathbb{R}^{N\times 2}$ and the state matrix $\mathbf{X} \in \mathbb{R}^{N\times 7}$ with $N=90702$.
Finally, the output vector $Y$, given by the values of $\mathbf{\ddot{\theta}}$, is computed using an acausal filter for our supervised model learning algorithm. Note that $\mathbf{\ddot{\theta}}$ are not needed online while controlling CMS and an acausal filter is important in order to reduce the delay and improve the accuracy of the signal to be learned. Finally, the data collected are divided by rings, and split into two to obtain 4 training and testing data sets (ring 5 is the target, so it is not considered in the learning algorithm). The time of use of the CMS needed by our learning algorithm is less then 25 minutes. To limit the computational complexity of the algorithms, the 4 training data sets are uniformly downsampled with coefficient $c_{DS}$, in order to obtain training sets of size $N_{tr}$ less than 5000 points; the size of the test set is denoted by $N_{test}$. During data collection, the ball naturally tends to stay more in the outermost ring, with the effect that the number of data per ring decreases going from ring 1 to ring 4 (see Table \ref{table:nRMSE_1step}). Each of the models described in Section~\ref{subsec:model_learning} and denoted by P, PI, NP, and SP, has been learned and tested with the data set for each ring and in the real CMS.
The quality of the learned models is evaluated in terms of nRMSE in training and test data for each ring and each model in Table~\ref{table:nRMSE_1step}.
Model P represents only an approximation of $\mathbf{\ddot{\theta}}$ and, as expected, its prediction performance is significantly worse than the other methods.  Similarly, PI performs poorly, however, it outperforms P, confirming the benefit of using the nonlinear behaviour of the state suggested by the physics as basis functions. The best methods are NP and SP, and we can notice that SP always outperforms NP. This shows the higher robustness of the semiparametric model due to its use of physical basis functions. 
Moreover, as the ring becomes smaller, the prediction performance deteriorates for all the models, and the gap in the performances of NP and SP between training and test increases when the amount of data available decreases, namely in ring 3 and 4. This agrees with theoretical expectations, and is likely due to the lower variability present in the data set. Finally, for comparison purposes, we added the performance of Random Forest regression (RF) \cite{breiman2001random}, a standard machine learning method based on a different paradigm than GPR. Table~\ref{table:nRMSE_1step} highlights that NP and SP exhibit better performance than RF. In Figure \ref{fig:nRMSE_learning_curve}, we compare the learning curves of the two best performing models (SP and NP) as the number of training data increases for ring 1. SP outperforms NP in terms of nRMSE at each evaluated $N_{tr}$, and achieves the same performance with half the data (note SP performance at $N_{tr}=2152$ and NP at $N_{tr}=3874$). 
After the evaluation of the prediction of acceleration, we are now interested on evaluating the n-steps-ahead prediction performance of the actual state variables $\theta$ and $\dot{\theta}$. This hard modelling task is fundamental for trajectory optimization control algorithms. 
Results are shown over a 40-step horizon, which is commensurate with the time needed for the ball to reach an opening, as we will see later in the experiments on the CME. From the test datasets, we randomly chose trajectories of length 40 steps that terminate in front of an opening (gate). The models are initialized with the initial conditions of these trajectories, and rolled out with the same control actions. 
In Figure~\ref{fig:rollout_performance}, the results of all the models in ring 1 are shown, evaluating the absolute errors in position $\vert \theta_k - \hat{\theta}_k\vert$ and velocity $\vert \dot{\theta}_k - \hat{\dot{\theta}}_k \vert$ at each prediction step $[1,\ldots,40]$, and reporting the mean over the 226 trajectories. This error index has been chosen for better interpretability of the control performance. 

\begin{table}[ht]
\footnotesize
\begin{tabular}{ c c c c c c c c c c c c c c c c}
\toprule
& \multicolumn{2}{c}{PI [rad]} & \multicolumn{2}{c}{NP [rad]} & \multicolumn{2}{c}{NPd [rad]} & \multicolumn{2}{c}{SP [rad]}\\
\midrule
r & n=20 & n=40 &n=20& n=40 & n=20 & n=40 & n=20 & n=40\\
\midrule
1 & 0.40 & 1.32 & 0.11 & 0.26 & 1.36 & 3.01 & 0.11 & 0.25\\
2 & 0.21 & 0.57 & 0.15 & 0.33 & 0.87 & 2.09 & 0.15 & 0.33\\
3 & 0.24 & 0.64 & 0.19 & 0.34 & 0.99 & 2.26 & 0.18 & 0.32\\
4 & 1.81 & 1.21 & 0.46 & 1.76 & 0.65 & 2.45 & 1.74 & 1.18\\
\bottomrule
\end{tabular}
\caption{Rollout performance for all the rings evaluated as the average absolute error in $\theta$ at 20 and 40 steps ahead.}
\label{table:rollout_performance}
\end{table}

In this comparison, we considered also the estimation of the discrete static map \eqref{eq:discrete_dynamics_general} using nonparametric GPR described in Section~\ref{subsubsec:nonparametric_model}, trained on the same data sets as all the other models and using the same kernel function~of NP. This model is denoted as Nonparametric discrete,~``NPd".
\begin{figure}
	\flushleft
	\includegraphics[width=\columnwidth]{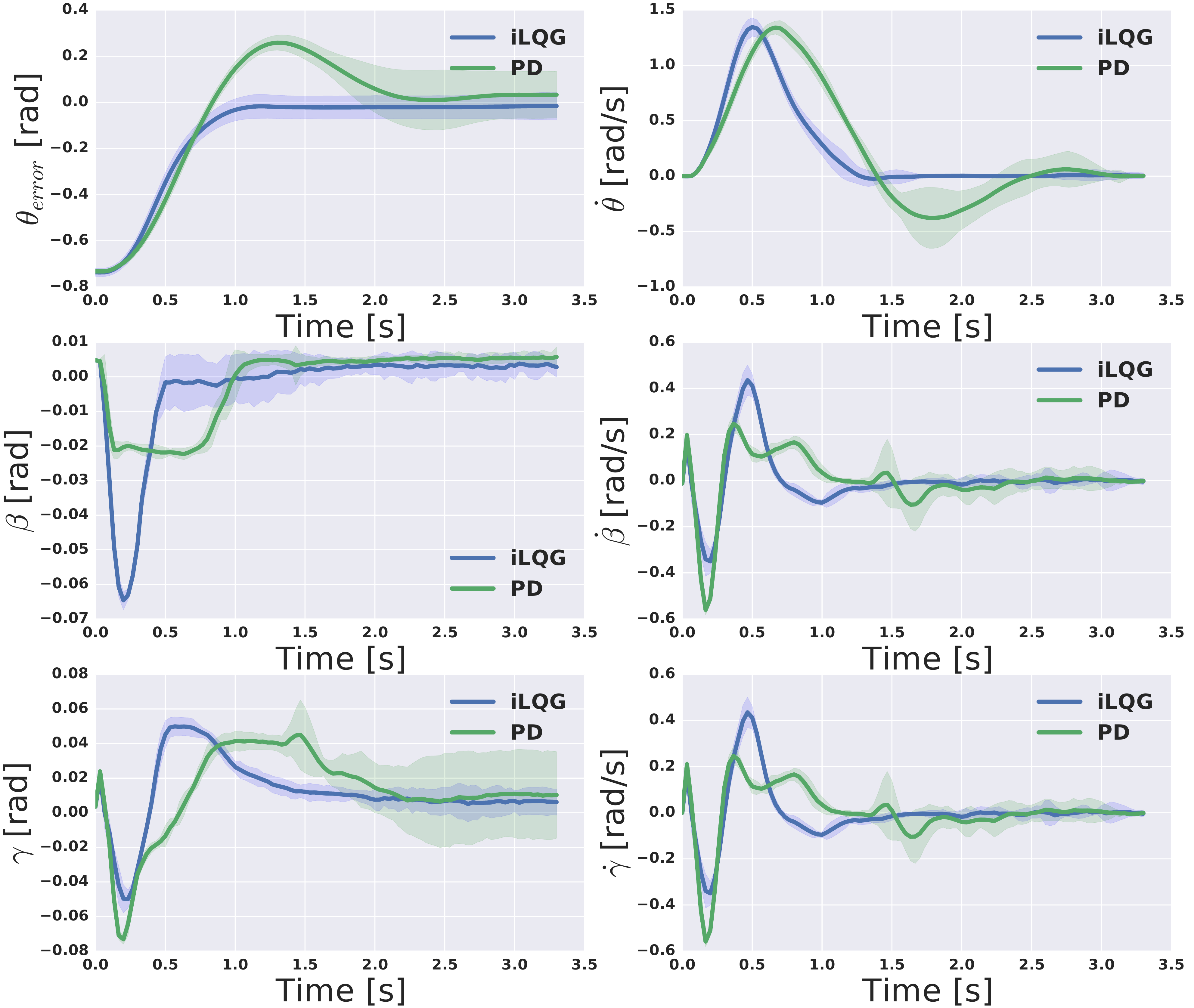}
	\caption{Comparison of the iLQG controller with PD in the outer-most ring of the maze over 10 runs. The steady error for both controllers is comparable to the size of the ball.}
	\label{fig:controller_comparison}
\vspace{-7 mm}
\end{figure}
All models diverge rapidly from the ground truth, except for SP and NP, which achieve competitive performance. Model SP consistently outperforms all the others with an average error in position of 14 degrees after 40 steps. The confidence interval at 95$\%$ for model SP and NP is also shown, from which we can see that there are trajectories where contacts and bounces make the dynamics hard to predict. Finally, in Table~\ref{table:rollout_performance}, the absolute error $\vert \dot{\theta}_k - \hat{\dot{\theta}}_k \vert$ is computed at 20 and 40 steps ahead for all the models (except P which performs poorly) in all the rings. Consistently, model SP and NP outperform the others in all of the rings. The significantly higher performance of NP w.r.t. NPd shows the benefit of formulating the learning problem as a dynamical system, as in \eqref{eq:continuous_dynamics_ball}, and not as a static map, as in \eqref{eq:discrete_dynamics_general}. The performance in ring 4 deteriorates significantly, even for model SP, because as explained before, the data collected in this ring are much less and of lower quality in comparison to the other rings. This problem will become clear while controlling the ball in CME. 
Next, we present results for trajectory optimization using iterative LQG~\cite{todorov2005, tassa2012}.  To demonstrate correctness of the models that were learned in the last section, we first show several motions to a target point in the outermost ring of the maze using the learned model. 
We use the cost function proposed by \cite{tassa2012}, which penalizes any deviation from the target state, as well as saturation of the control signal. 
The state cost is the smooth-abs function $\ell(x)=\sqrt{x^2+\alpha^2}-\alpha$, where $x$ represents state error and $\alpha$ is a parameter to control the smoothness around origin. The control cost is the function $\ell(u)=\nu^2(\cosh(u/\nu)-1)$ which penalizes input saturation, where $u$ is the control input and by varying $\nu$, we can limit the control signal to a particular volume of control space. 
We compare the results with a well-tuned PD controller. 
The error plots for 10 such trajectories, for a movement of $\pi/4$ radians in the outer-most ring, are shown in Figure~\ref{fig:controller_comparison}. We can see that the iLQG controller is behaving in a nearly time-optimal fashion using the model learned earlier. We repeat the same tests for all the rings for a movement of $\pi/4$ radians. With iLQG, we achieve a settling time of $1.5s$ and $2.73s$, and steady state error of $0.013$ and $0.05$ radians, for rings $2$ and $3$, respectively. With PD, we achieved $2.0s$ and $2.70s$ with error of $0.066$ and $0.012$ radians in rings $2$ and $3$, respectively. In the fourth ring, our model isn't very accurate, and, without feedback tracking, we were not able to regulate the position of the ball for very small movements. To solve the full problem of navigating to the inner-most goal state of the maze, we implement the trajectories computed for each of the rings with trajectory tracking using a PD controller. The radial transition was implemented when the ball was in front of an opening using the control signal described in Section~\ref{subsec:radial_movement}. This resulted in a reasonably good performance. 
However, as the trajectories are local, there were some failure runs due to contacts with the walls in the CME (the contact dynamics, such as bouncing off the walls, were not included in the model). 
We tried $10$ runs for the full trajectory, where $7$ runs were successful with an average time of $7.71s$, best time of $6.2s$, and worst time of~$10.13s$. 
%
\section{Conclusion and Discussion}
\label{sec:conclusion}
In this paper, the problem of learning how to control a ball in a CME is presented. Our approach offers an initial solution to navigate the ball to the goal-state of the CME. The goal of reaching the final target state is decomposed into a sequence of micro-goals which consist of navigating in each ring, each of which is solved by a trajectory optimization algorithm using the learned models. Accurate semiparametric GP models are learned to describe the acceleration function of the dynamical system.
While the current controller can navigate the ball to the goal state in the CME reasonably well, it is still local, and not robust to changes in initial conditions and other disturbances due to contacts. We are currently working on learning a full policy using a guided policy search~\cite{LK13} approach, where we can optimize trajectory and policy simultaneously. Moreover, we are investigating on how to apply GP online learning techniques  \cite{Lawrence:2002,Snelson06sparsegaussian,AE-MY-JH:11,2018arXiv180905074R,romeres2016line,gijsberts2011incremental} to improve the model while iteratively improving the policy. Finally, we are interested in composing the prior physics knowledge with a neural network learning framework and compare it with the proposed semiparametric GP models~\cite{2018arXiv180803246A}. 
Apart from this, the proposed CMS could be used to study several other interesting problems. For example, it would be interesting to see how the models learned for a single ball could be leveraged to compose accurate motion models for more than one ball in the environment. 

The CMS is suitable to study several problems in RL and robot learning presenting several challenges such as nonlinear and fast dynamics, contacts and thus it could possibly be used as a benchmark system for its ease of~reproducibility.


\bibliographystyle{ieeetr}
\bibliography{main.bib}  

\begin{thebibliography}{10}

\bibitem{GS02}
G.~Gilardi and I.~Sharf, ``Literature survey of contact dynamics modelling,''
  {\em Mechanism and machine theory}, vol.~37, no.~10, pp.~1213--1239, 2002.

\bibitem{kloss2017combining}
A.~Kloss, S.~Schaal, and J.~Bohg, ``Combining learned and analytical models for
  predicting action effects,'' {\em arXiv preprint arXiv:1710.04102}, 2017.

\bibitem{fazeli2017learning}
N.~Fazeli, S.~Zapolsky, E.~Drumwright, and A.~Rodriguez, ``Learning
  data-efficient rigid-body contact models: Case study of planar impact,'' in
  {\em Conference on Robot Learning}, pp.~388--397, 2017.

\bibitem{ajay2018augmenting}
A.~Ajay, J.~Wu, N.~Fazeli, M.~Bauza, L.~P. Kaelbling, J.~B. Tenenbaum, and
  A.~Rodriguez, ``Augmenting physical simulators with stochastic neural
  networks: Case study of planar pushing and bouncing,'' {\em arXiv preprint
  arXiv:1808.03246}, 2018.

\bibitem{silver2017}
D.~Silver, J.~Schrittwieser, K.~Simonyan, I.~Antonoglou, A.~Huang, A.~Guez,
  T.~Hubert, L.~Baker, M.~Lai, A.~Bolton, {\em et~al.}, ``Mastering the game of
  go without human knowledge,'' {\em Nature}, vol.~550, no.~7676, p.~354, 2017.

\bibitem{mnih2015}
V.~Mnih, K.~Kavukcuoglu, D.~Silver, A.~A. Rusu, J.~Veness, M.~G. Bellemare,
  A.~Graves, M.~Riedmiller, A.~K. Fidjeland, G.~Ostrovski, {\em et~al.},
  ``Human-level control through deep reinforcement learning,'' {\em Nature},
  vol.~518, no.~7540, p.~529, 2015.

\bibitem{levine2016}
S.~Levine, C.~Finn, T.~Darrell, and P.~Abbeel, ``End-to-end training of deep
  visuomotor policies,'' {\em The Journal of Machine Learning Research},
  vol.~17, no.~1, pp.~1334--1373, 2016.

\bibitem{SB98}
R.~S. Sutton and A.~G. Barto, {\em Reinforcement learning: An introduction},
  vol.~1.
\newblock MIT press Cambridge, 1998.

\bibitem{KBP13}
J.~Kober, J.~A. Bagnell, and J.~Peters, ``Reinforcement learning in robotics: A
  survey,'' {\em The International Journal of Robotics Research}, vol.~32,
  no.~11, pp.~1238--1274, 2013.

\bibitem{DNP13}
M.~P. Deisenroth, G.~Neumann, J.~Peters, {\em et~al.}, ``A survey on policy
  search for robotics,'' {\em Foundations and Trends{\textregistered} in
  Robotics}, vol.~2, no.~1--2, pp.~1--142, 2013.

\bibitem{ASM08}
A.~Antos, C.~Szepesv{\'a}ri, and R.~Munos, ``Fitted {Q}-iteration in continuous
  action-space {MDP}s,'' in {\em Advances in neural information processing
  systems}, pp.~9--16, 2008.

\bibitem{DFR15}
M.~P. Deisenroth, D.~Fox, and C.~E. Rasmussen, ``Gaussian processes for
  data-efficient learning in robotics and control,'' {\em IEEE Transactions on
  Pattern Analysis and Machine Intelligence}, vol.~37, no.~2, pp.~408--423,
  2015.

\bibitem{DR11}
M.~Deisenroth and C.~E. Rasmussen, ``{PILCO}: A model-based and data-efficient
  approach to policy search,'' in {\em Proceedings of the 28th International
  Conference on machine learning (ICML-11)}, pp.~465--472, 2011.

\bibitem{LK13}
S.~Levine and V.~Koltun, ``Guided policy search,'' in {\em Proceedings of the
  30th International Conference on Machine Learning (ICML-13)}, pp.~1--9, 2013.

\bibitem{boedecker2014approximate}
J.~Boedecker, J.~T. Springenberg, J.~W{\"u}lfing, and M.~Riedmiller,
  ``Approximate real-time optimal control based on sparse gaussian process
  models,'' in {\em Adaptive Dynamic Programming and Reinforcement Learning
  (ADPRL), 2014 IEEE Symposium on}, pp.~1--8, IEEE, 2014.

\bibitem{kamthe2017data}
S.~Kamthe and M.~P. Deisenroth, ``Data-efficient reinforcement learning with
  probabilistic model predictive control,'' {\em arXiv preprint
  arXiv:1706.06491}, 2017.

\bibitem{lee2017gp}
G.~Lee, S.~S. Srinivasa, and M.~T. Mason, ``G{P}-{ILQG}: Data-driven robust
  optimal control for uncertain nonlinear dynamical systems,'' {\em arXiv
  preprint arXiv:1705.05344}, 2017.

\bibitem{St-Atkeson-06}
M.~Stolle and C.~Atkeson, ``Policies based on trajectory libraries,'' in {\em
  Proceedings of the International Conference on Robotics and Automation},
  January 2006.

\bibitem{zucker2012}
M.~Zucker and J.~A. Bagnell, ``Reinforcement planning: {RL} for optimal
  planners,'' in {\em Robotics and Automation (ICRA), 2012 IEEE International
  Conference on}, pp.~1850--1855, IEEE, 2012.

\bibitem{bentivegna2001}
D.~C. Bentivegna and C.~G. Atkeson, ``Learning from observation using
  primitives,'' in {\em Robotics and Automation, 2001. Proceedings 2001 ICRA.
  IEEE International Conference on}, vol.~2, pp.~1988--1993, IEEE, 2001.

\bibitem{2018arXiv180904720V}
J.~{van Baar}, A.~{Sullivan}, R.~{Cordorel}, D.~{Jha}, D.~{Romeres}, and
  D.~{Nikovski}, ``{Sim-to-Real Transfer Learning using Robustified Controllers
  in Robotic Tasks involving Complex Dynamics},'' {\em ArXiv e-prints}, Sept.
  2018.

\bibitem{romeres2016onlineIcub}
D.~Romeres, M.~Zorzi, R.~Camoriano, and A.~Chiuso, ``Online semi-parametric
  learning for inverse dynamics modeling,'' in {\em Conference on Decision and
  Control (CDC), 2016 IEEE 55th}, pp.~2945--2950, IEEE, 2016.

\bibitem{wu2012semi}
T.~Wu and J.~Movellan, ``Semi-parametric {G}aussian process for robot system
  identification,'' in {\em IEEE/RSJ International Conference on Intelligent
  Robots and Systems (IROS)}, pp.~725--731, 2012.

\bibitem{ICRA2010NguyenTuong_62320}
D.~Nguyen-Tuong and J.~Peters, ``Using model knowledge for learning inverse
  dynamics,'' in {\em IEEE International Conference on Robotics and
  Automation}, 2010.

\bibitem{Rasmussen}
C.~Rasmussen and C.~Williams, {\em {G}aussian Processes for Machine Learning}.
\newblock The MIT Press, 2006.

\bibitem{2018arXiv180905074R}
D.~{Romeres}, M.~{Zorzi}, R.~{Camoriano}, S.~{Traversaro}, and A.~{Chiuso},
  ``{Derivative-free online learning of inverse dynamics models},'' {\em ArXiv
  e-prints}, Sept. 2018.

\bibitem{Nguyen-Tuong2011}
D.~Nguyen-Tuong and J.~Peters, ``Model learning for robot control: a survey,''
  {\em Cognitive Processing}, vol.~12, no.~4, pp.~319--340, 2011.

\bibitem{todorov2005}
E.~Todorov and W.~Li, ``A generalized iterative {LQG} method for
  locally-optimal feedback control of constrained nonlinear stochastic
  systems,'' in {\em American Control Conference, 2005. Proceedings of the
  2005}, pp.~300--306, IEEE, 2005.

\bibitem{tassa2012}
Y.~Tassa, T.~Erez, and E.~Todorov, ``Synthesis and stabilization of complex
  behaviors through online trajectory optimization,'' in {\em Intelligent
  Robots and Systems (IROS), 2012 IEEE/RSJ International Conference on},
  pp.~4906--4913, IEEE, 2012.

\bibitem{ROS09}
M.~Quigley, K.~Conley, B.~P. Gerkey, J.~Faust, T.~Foote, J.~Leibs, R.~Wheeler,
  and A.~Y. Ng, ``{ROS}: an open-source robot operating system,'' in {\em ICRA
  Workshop on Open Source Software}, 2009.

\bibitem{kroemer2014}
O.~Kroemer, H.~Van~Hoof, G.~Neumann, and J.~Peters, ``Learning to predict
  phases of manipulation tasks as hidden states,'' in {\em Robotics and
  Automation (ICRA), 2014 IEEE International Conference on}, pp.~4009--4014,
  IEEE, 2014.

\bibitem{lioutikov2015}
R.~Lioutikov, G.~Neumann, G.~Maeda, and J.~Peters, ``Probabilistic segmentation
  applied to an assembly task,'' in {\em Humanoid Robots (Humanoids), 2015
  IEEE-RAS 15th International Conference on}, pp.~533--540, IEEE, 2015.

\bibitem{kroemer2015}
O.~Kroemer, C.~Daniel, G.~Neumann, H.~Van~Hoof, and J.~Peters, ``Towards
  learning hierarchical skills for multi-phase manipulation tasks,'' in {\em
  Robotics and Automation (ICRA), 2015 IEEE International Conference on},
  pp.~1503--1510, IEEE, 2015.

\bibitem{anjali2016implementation}
T.~Anjali and S.~S. Mathew, ``Implementation of optimal control for ball and
  beam system,'' in {\em Emerging Technological Trends (ICETT), International
  Conference on}, pp.~1--5, IEEE, 2016.

\bibitem{campos2004optimal}
D.~U. Campos-Delgado, ``Optimal tracking controller for ball and beam system,''
  2004.

\bibitem{Ljung:99}
L.~Ljung, {\em System Identification - Theory for the User}.
\newblock Upper Saddle River, N.J.: Prentice-Hall, 2nd~ed., 1999.

\bibitem{siciliano2010robotics}
B.~Siciliano, L.~Sciavicco, L.~Villani, and G.~Oriolo, {\em Robotics:
  modelling, planning and control}.
\newblock Springer Science \& Business Media, 2010.

\bibitem{bullo2004}
F.~Bullo and A.~D. Lewis, {\em Geometric control of mechanical systems:
  modeling, analysis, and design for simple mechanical control systems},
  vol.~49.
\newblock Springer Science \& Business Media, 2004.

\bibitem{hollerbach2008model}
J.~Hollerbach, W.~Khalil, and M.~Gautier, ``Model identification,'' in {\em
  Springer Handbook of Robotics}, pp.~321--344, Springer, 2008.

\bibitem{breiman2001random}
L.~Breiman, ``Random forests,'' {\em Machine learning}, vol.~45, no.~1,
  pp.~5--32, 2001.

\bibitem{Lawrence:2002}
N.~Lawrence, M.~Seeger, and R.~Herbrich, ``Fast sparse {G}aussian process
  methods: The informative vector machine,'' in {\em Proceedings of the 15th
  International Conference on Neural Information Processing Systems}, NIPS'02,
  (Cambridge, MA, USA), pp.~625--632, MIT Press, 2002.

\bibitem{Snelson06sparsegaussian}
E.~Snelson and Z.~Ghahramani, ``Sparse gaussian processes using
  pseudo-inputs,'' in {\em Advances in Neural Information Processing Systems},
  pp.~1257--1264, MIT press, 2006.

\bibitem{AE-MY-JH:11}
A.~Ranganathan, M.~H. Yang, and J.~Ho, ``Online sparse {G}aussian process
  regression and its applications,'' {\em IEEE Transactions on Image
  Processing}, vol.~20, pp.~391--404, Feb 2011.

\bibitem{romeres2016line}
D.~Romeres, G.~Prando, G.~Pillonetto, and A.~Chiuso, ``On-line bayesian system
  identification,'' in {\em Control Conference (ECC), 2016 European},
  pp.~1359--1364, IEEE, 2016.

\bibitem{gijsberts2011incremental}
A.~Gijsberts and G.~Metta, ``Incremental learning of robot dynamics using
  random features,'' in {\em IEEE International Conference on Robotics and
  Automation (ICRA)}, pp.~951--956, 2011.

\bibitem{2018arXiv180803246A}
A.~{Ajay}, J.~{Wu}, N.~{Fazeli}, M.~{Bauza}, L.~P. {Kaelbling}, J.~B.
  {Tenenbaum}, and A.~{Rodriguez}, ``{Augmenting Physical Simulators with
  Stochastic Neural Networks: Case Study of Planar Pushing and Bouncing},''
  {\em ArXiv e-prints}, Aug. 2018.

\end{thebibliography}

\end{document}